\documentclass[runningheads]{llncs}
\usepackage[T1]{fontenc}
\usepackage{amsmath,amsfonts,amssymb}
\usepackage{graphicx}
 \usepackage{epstopdf}
 \usepackage{booktabs} 
\usepackage{color}
\usepackage{amssymb}

\usepackage{todonotes}

\begin{document}
\title{\textbf{D}ense \textbf{T}ransformer based \textbf{E}nhanced \textbf{C}oding Network for Unsupervised Metal Artifact Reduction}
\titlerunning{DTEC-Net for Unsupervised Metal Artifact Reduction}
\author{Wangduo Xie 
\and
Matthew B.\ Blaschko} 
\authorrunning{W.\ Xie and M.\ B.\ Blaschko}
\institute{Center for Processing Speech and Images, Department of ESAT,\\ KU Leuven, Leuven, Belgium\\
\email{firstname.lastname@esat.kuleuven.be}}
\maketitle              
\begin{abstract}
 CT images corrupted by metal artifacts have serious negative effects on clinical diagnosis. Considering the difficulty of collecting paired data with ground truth in clinical settings, unsupervised methods for metal artifact reduction are of high interest. However, it is difficult for previous unsupervised methods to retain structural information from CT images while handling the non-local characteristics of metal artifacts. To address these challenges, we proposed a novel \textit{\textbf{D}ense \textbf{T}ransformer based \textbf{E}nhanced \textbf{C}oding \textbf{Net}work} (\textbf{DTEC-Net}) for unsupervised metal artifact reduction. Specifically, we introduce a Hierarchical Disentangling Encoder, supported by the high-order dense process, and transformer to obtain densely encoded sequences with long-range correspondence. Then, we present a second-order disentanglement method to improve the dense sequence's decoding process.  Extensive experiments and model discussions illustrate DTEC-Net's effectiveness, which outperforms the previous state-of-the-art methods on a benchmark dataset, and greatly reduces metal artifacts while restoring richer texture details.
\end{abstract}
\section{Introduction}
CT technology can recover the internal details of the human body in a non-invasive way and has been widely used in clinical practice. However, if there is metal in the tissue, metal artifacts (MA) will appear in the reconstructed CT image, which will corrupt the image and affect the medical diagnosis \cite{barrett2004artifacts,lemmens2008suppression}.

In light of the clinical need for MA reduction, various traditional methods \cite{kalender1987reduction,lemmens2008suppression,meyer2010normalized,zhang2016iterative} have been proposed to solve the problem by using interpolation  and iterative optimization. As machine learning research increasingly impacts medical imaging, deep learning  based methods have been proposed for MA reduction.  Specifically, these methods can be roughly divided into supervised and unsupervised categories according to the degree of supervision.  In the supervised category, the methods \cite{lin2019dudonet,wang2023indudonet+,wang2021dual,yu2020deep} based on the dual domain (sinogram and image domains) can achieve good performance for MA reduction. However, supervised learning methods are hindered by the lack of large-scale real-world data pairs consisting of "images with MA" and "images without MA" representing the same region. The lack of such data can lead algorithms trained on synthetic data to over-fit simulated data pairs, resulting in difficulties in generalizing to clinical settings \cite{lyu2021u}. Furthermore, although sinogram data can bring additional information, it is difficult to collect in realistic settings \cite{liao2019adn,wang2022orientation}. Therefore,  \textit{unsupervised methods based only on the image domain} are strongly needed in practice.

For unsupervised methods in the image domain, Liao \textit{et al}.\  \cite{liao2019adn} used Generative Adversarial Networks (GANs) \cite{goodfellow2020generative} to disentangle the MA from the underlying clean structure of the artifact-affected image in latent space by using unpaired data with and without MA.
Although the method can separate the artifact component in the latent space, the features from the latent space can't represent rich low-level information of the original input. Further, it's also hard for the encoder to represent long-range correspondence across different regions. Accordingly, the restored image loses texture details and can't retain structure from the CT image. In the same unsupervised setting, Lyu \textit{et al}.\ \cite{lyu2021u} directly separate the MA component and clean structure in image space using a CycleGAN-based method \cite{zhu2017unpaired}. Although implementation in the image space makes it possible to construct dual constraints, directly operating in the image space affects the algorithm's performance upper limit, because it is difficult to encode in the image space as much low-level information as the feature space.

Considering the importance of low-level features in the latent space for generating the artifact-free component, we propose a 
novel \textit{Dense Transformer based Enhanced Coding Network}(DTEC-Net) for unsupervised metal artifact reduction, which can obtain low-level features with hierarchical information and map them to a clean image space through adversarial training. DTEC-Net contains our developed Hierarchical Disentangling Encoder (HDE), which utilizes long-range correspondences obtained by a lightweight transformer and a high-order dense process to produce the enhanced coded sequence. To ease the burden of decoding the sequence, we also propose a second-order disentanglement method to finish the sequence decomposition. Extensive empirical results show that our method can not only reduce the MA greatly and generate high-quality images, but also surpasses the competing
unsupervised approaches. 
\begin{figure}[t]
\centering
\includegraphics[width=0.95\textwidth]{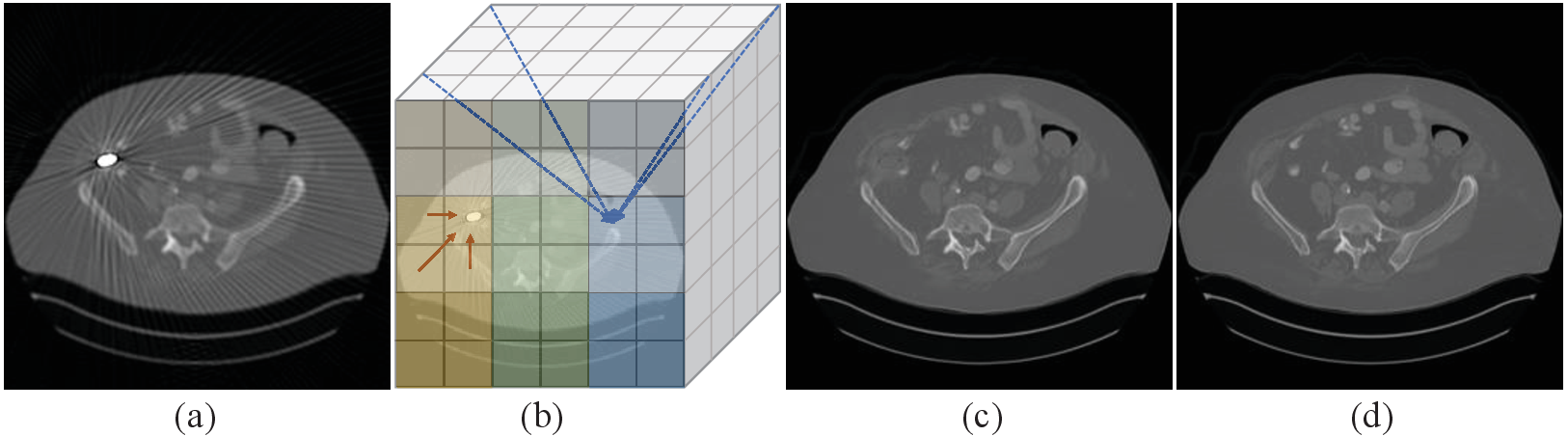}
\caption{(a) CT image with metal artifacts. (b) Blue/Orange arrows: Reuse of low-level features/Long-range correspondence. (c) Output of our DTEC-Net. (d) Ground truth.}
\label{begin}
\end{figure}

\section{Methodology}

We design a Hierarchical Disentangling Encoder(HDE) that can capture low-level sequences and enable high-performance restoration. Moreover, to reduce the burden of the decoder group brought by the complicated sequences, we propose a second-order disentanglement mechanism. The intuition is shown in Fig.~\ref{begin}.

\subsection{Hierarchical Disentangling Encoder (HDE)}\label{sec:HDE}
\begin{figure}[t]
\centering
\includegraphics[width=0.95\textwidth]{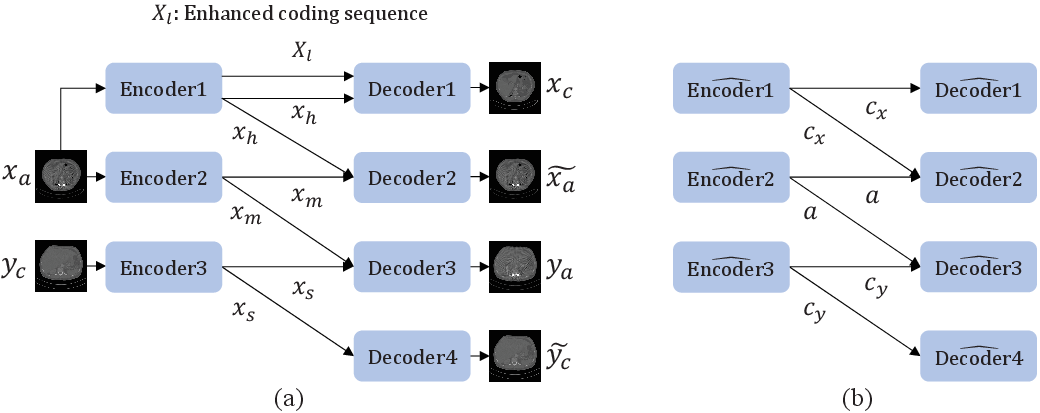}
\caption{(a) Generator of DTEC-Net. $x_{a}$: input with artifacts. $y_{c}$:  unpaired input without artifacts. $X_{l}$ and $x_{h}$ are defined in the section 2.1. $x_{m}$: the MA parts in latent space. $x_{s}$: the overall “structural part” in latent space. $x_{c}$ (or $y_{a}$): the output after removing (or adding) the artifacts. $\widetilde{x_{a}}$ (or $\widetilde{y_{c}}$): the output of the identity map.  (b) Generator of ADN \cite{liao2019adn}. The data relationship is shown in \cite{liao2019adn}. In addition to the difference in disentanglement, DTEC-Net and ADN also have different inner structures.}
\label{generator}
\end{figure}

\begin{figure}[htpb]
\centering
\includegraphics[width=0.9\textwidth]{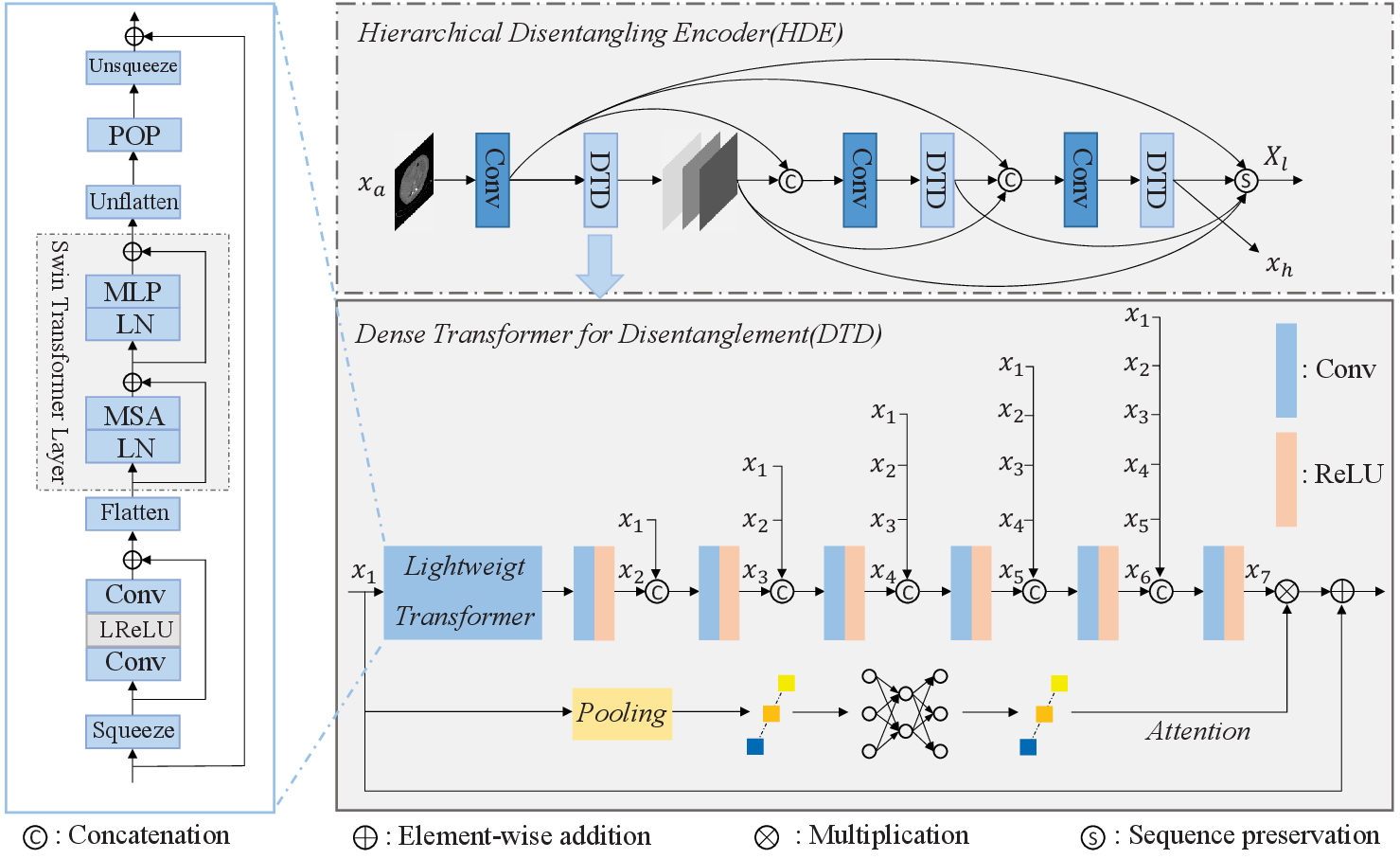}
\caption{The architecture of Hierarchical Disentangling Encoder(HDE).}
\label{big_image_2}
\end{figure}

As shown in Fig.~\ref{generator}(a), the generator of DTEC-Net consists of three encoders and four decoders. 
We design the HDE to play the role of Encoder1 for enhanced coding. Specifically, for the HDE's input image $x_{a}\in R^{1 \times H \times W}$ with MA, HDE first uses a convolution for the preliminary feature extraction and produces a high-dimensional tensor $x_{l_0}$ with $c$ channels. Then, $x_{l_0}$ will be encoded by three \textit{Dense Transformers for Disentanglement} (DTDs) in a first-order reuse manner \cite{huang2017densely,zhang2018residual}. Specifically, the output $x_{l_i}$ of the $i$th DTD can be characterized as:
\begin{equation}
x_{l_i}=\left\{\begin{array}{l}
f_{\operatorname{DTD}_i}(f_{\operatorname{s-hde}}(\operatorname{cat}(x_{l_{i-1}},x_{l_{i-2}}...,,x_{l_0}))), i=2,...,N.  \\
f_{\operatorname{DTD}_i}(x_{l_{i-1}}), i=1. \\
\end{array} \right.
\label{HDE}
\end{equation}
In Equation~\eqref{HDE}, $f_{\operatorname{s-hde}}$ represents the channel compression of the concatenation of multiple DTDs' outputs, and $N$ represents the total number of DTDs in the HDE. As shown in Fig.~\ref{big_image_2}, HDE can obtain the hierarchical information sequence $X_{l}\triangleq\{x_{l_0},...x_{l_{N}}\}$ and high-level semantic features $x_{h}\triangleq x_{l_{N}}$.

As shown in Fig.~\ref{generator}(b), Encoder1 of ADN cannot characterize upstream low-level information, and results in limited performance. By using HDE, the upstream of the DTEC-Net's Encoder1 can represent rich low-level information, and be encoded in the efficient way described in Equation~\eqref{HDE}. After 
generating the enhanced coding sequences $X_{l}$ with long-range correspondence and densely reused information, DTEC-Net can decode it back to the clean image domain by using the proposed second-order disentanglement for MA reduction, which reduces the decoder group's burden to a large extent.

\subsection{Dense Transformer for Disentanglement (DTD)}
In addition to the first-order feature multiplexing given in Equation~\eqref{HDE}, HDE also uses the DTD to enable second-order feature reuse. The relationship between HDE and DTD is shown in Fig.~\ref{big_image_2}. 
Inspired by \cite{liang2021swinir,zhang2022accurate}, DTD first uses a lightweight transformer based on the Swin transformer \cite{liang2021swinir,liu2021swin} to represent content-based information with long-range correspondence inside of every partition window. It then performs in-depth extraction and second-order reuse.

Specifically, the input $x_{1}\in R^{C \times H \times W}$ of the DTD will be processed sequentially by the lightweight transformer and groups of convolutions in the form of second-order dense connections.
The output $x_{j+1}$ of the $j$th convolution with ReLU, which is connected in a second-order dense pattern, can be expressed as: 
\begin{equation}
x_{j+1}=\Big\{
\begin{array}{l}
f_{c_j}(\operatorname{cat}(x_1, x_2,..., x_j)), j=2,3,...,J.\\
f_{c_j}(f_{\operatorname{transformer-light}}(x_{j})), j=1. \\
\end{array}  
\label{detail_DTD}
\end{equation}
In Equation~\eqref{detail_DTD}, $f_{c_j}$ indicates the $j$th convolution with ReLU after the lightweight transformer, and the $J$ indicates the total number of convolutions after the lightweight transformer and is empirically set to six. The dense connection method can effectively reuse low-level features \cite{zhang2018image,zhang2018residual} so that the latent space including these type of features will help the decoder to restore clean images without metal artifacts. Because the low-level information on different channels has different importance to the final restoration task, we use the channel attention mechanism \cite{hu2018squeeze} to filter the output of the final convolution layer:
\begin{equation}\begin{array}{l}
x_{\operatorname{out}}=x_{J+1} \odot f_{\operatorname{MLP}}(f_{\operatorname{pooling}}(x_1))+x_1,
\end{array} 
\end{equation}
where $\odot$ represents the Hadamard product, $f_{\operatorname{MLP}}$ indicates a multi-layer perceptron with only one hidden layer, and $f_{\operatorname{pooling}}$ represents global pooling.

Because the transformer usually requires a large amount of data for training and CT image datasets are usually smaller than those for natural images, we do lightweight processing for the Swin transformer. Specifically, for an input tensor $x\in R^{C \times H \times W}$ of the lightweight transformer, the number of channels will be reduced from $C$ to $C_{\operatorname{in}}$ to lighten the burden of the attention matrix. Then, a residual block is employed to extract information with low redundancy. 

After completing lightweight handling, the tensor will first be partitioned into multiple local windows and flattened to $x_{\operatorname{in}}\in R^{(\frac{HW}{P^2})\times {P^2} \times C_{in}}$ according the pre-operation \cite{liang2021swinir,liu2021swin} of the Swin transformer. $P\times P$ represents the window size for partitioning as shown in Fig.~\ref{begin}(b). Then, the attention matrix belonging to the $i$th window can be calculated by pairwise multiplication between converted vectors in $S_{i}\triangleq \{x_{\operatorname{in}}(i, j, :)|j=0, 1, ..., P^2-1\}$. Specifically, by using a linear map from $R^{C_{\operatorname{in}}}$ to $R^{C_{a}}$ for every vector in $S_{i}$, the query key and value: $Q$, $K$, $V$ $\in  R^{{P^2}\times C_{a}}$ can be derived. Afterwards, the attention matrix for each window can be obtained by the following formula \cite{liu2021swin}:
\begin{equation}\begin{array}{l}
\operatorname{Attention}(Q, K, V)=\operatorname{SoftMax}(Q K^T / \sqrt{C_{a}}) V.
\end{array}
\end{equation}
In actual operation, we use window-based multi-head attention (MSA) \cite{liang2021swinir,liu2021swin,vaswani2017attention} to replace the single-head attention because of the performance improvement \cite{liang2021swinir}. The output of the Swin transformer layer will be unflattened and operated by post processing (POP) which consists of a classic convolution and layer norm (LN) with flatten and unflatten operations. After POP, the lightweight tensor with fewer channels will be unsqueezed to expand the channels, and finally added to the original input $x$ in the form of residuals.

\subsection{Second-order Disentanglement for MA Reduction (SOD-MAR)}\label{sec:SOD-MAR}
As mentioned in Section~\ref{sec:HDE}, $X_{l}$ represents the hierarchical sequence and facilitates the generator's representation. However, $X_{l}$ needs to be decoded by a high-capacity decoder to match the encoder. Considering that Decoder2 does not directly participate in the restoration branch and already loaded up the complicated artifact part $x_{m}$ in traditional first-order disentanglement learning \cite{liao2019adn}, to reduce the burden of the decoder group, we propose and analyze SOD-MAR.

\begin{figure}[t]
\centering
\includegraphics[width=\textwidth]{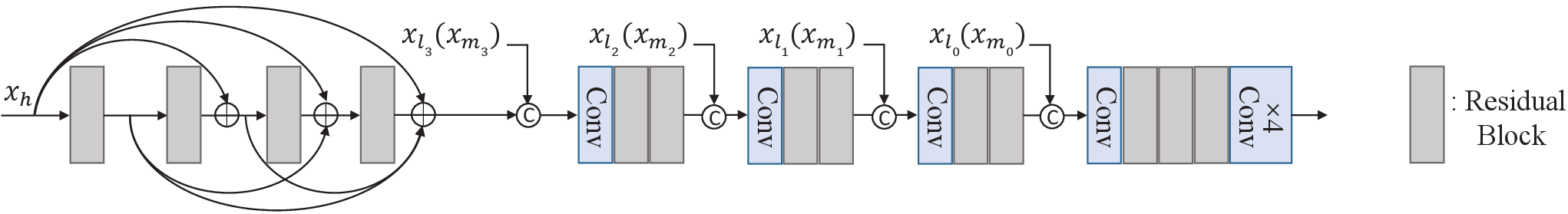}
\caption{The architecture of Decoder1 (Decoder2). ($x_{m_*}$) represents the Decoder2 case.}
\label{decoder_1}
\end{figure}

Specifically, Decoder2 of DTEC-Net doesn't decode sequence $X_{l}$, it only decodes the combination of second-order disentangled information $x_{h} \in X_{l}$ and the latent feature $x_{m}$ representing the artifact parts shown in Fig.~\ref{generator}(a). In order to complete the process, Decoder2 uses the structure shown in Fig.~\ref{decoder_1} to finish the decoding step, which is also used by Decoder1 to decode the sequence $X_{l}$.

Moreover, we don't only map the $x_{h}$ into Decoder1 and Decoder2 while dropping the $X_{l}$$\setminus$\{$x_{h}$\} to implement the burden reduction, because the low-level information in $X_{l}$$\setminus$\{$x_{h}$\} is vital for restoring artifact-free images. Furthermore, $x_{h}$ will be disturbed by noise from the approaching target $x_{a}$ of Decoder2 while information $X_{l}\setminus\{ x_{h} \}$ upstream from the HDE can counteract the noise disturbance to a certain extent. The reason behind the counteraction is that the update to upstream parameters is not as large as that of the downstream parameters.

\subsection{Loss function}

Following \cite{liao2019adn}, we use discriminators $D_{0}$, $D_{1}$ to constrain the output $x_c$ and $y_a$: 
\begin{equation}\begin{array}{l}
L_{adv}=\mathbb{E}[\log(1-D_0(x_c))+\log(D_0(y_c))]+\mathbb{E}[\log(D_1(x_a))+\log(1-D_1(y_a))].
\end{array}
\end{equation}
The above $x_a$, $y_c$ represent the input as shown in Fig. 2(a). Following \cite{liao2019adn}, we use the reconstruction loss $L_{rec}$ to constrain the identity map, and also use the artifact consistency loss $L_{art}$ and self-reduction loss $L_{self}$ to control the optimization process. The coefficients for each of these losses are set as in \cite{liao2019adn}.

\section{Empirical Results}
 \textbf{Synthesized DeepLesion Dataset} Following \cite{liao2019adn}, we randomly select 4186 images from DeepLesion \cite{yan2018deeplesion} and 100 metal templates \cite{zhang2018convolutional} to build a dataset. The simulation is consistent with \cite{liao2019adn,zhang2018convolutional}. For training, we randomly select 3986 images from DeepLesion combined with 90 metal templates for simulation. The  3986 images will be divided to two disjoint image sets with and without MA after simulation. Then a random combination can form the physically unpaired data with and without MA in the training process. Besides, another 200 images combined with the remaining 10 metal templates are used for the testing process.

\noindent
\textbf{Real Clinic Dataset} We randomly combine 6165 artifacts-affected images and 20729 artifacts-free images from SpineWeb$\footnote{spineweb.digitalimaginggroup.ca}$ \cite{liao2019adn} for training, and 105 artifacts-affected images from SpineWeb for testing.

\noindent
\textbf{Implementation Details} We use peak signal-to-noise ratio (PSNR) and structural similarity index (SSIM) to measure performance. We use mean squared error (MSE) only for measuring ablation experiments. For Synthesized DeepLesion dataset (and Real Clinic dataset), we set the batch size to 2 (and 2) and trained the network for 77 (and 60) epochs using the Adam optimizer. Our DTEC-Net was implemented in Pytorch using an Nvidia Tesla P100. 
\begin{table}[h]
	\begin{center}
\caption{Ablation study on Synthesized DeepLesion under different settings. $\uparrow$: Higher value is better; $\downarrow$: Lower value is better. The best values are in \textbf{bold}.}
\label{ta1}
\setlength{\tabcolsep}{3mm}{
\begin{tabular}{cccc}
\toprule
 Model & PSNR$\uparrow$ & SSIM$\uparrow$ & MSE$\downarrow$\\
\midrule
HDE with one DTD  & 34.46 & 0.937 & 27.12\\
HDE with two DTD  & 34.71 & 0.938 & 24.96\\
HDE with three DTD (only Transformer) & 34.31 & 0.936 &27.40\\
HDE with three DTD (without SOD-MAR) & 34.91 & 0.940 & 24.36\\
HDE with three DTD (with SOD-MAR, Ours) &  \textbf{35.11} &  \textbf{0.941} &\textbf{22.89}\\
\bottomrule
\end{tabular}
}
\end{center}
\end{table}

\subsection{Ablation Study}
To verify the effectiveness of the proposed methods, ablation experiments were carried out on Synthesized DeepLesion. The results are shown in Table \ref{ta1}.

\noindent
\textbf{The impact of DTD in HDE} 
In this experiment, we change the encoding ability of HDE by changing the number of DTDs. We first use only one DTD to build the HDE, then the PSNR is 0.65dB lower than our DTEC-Net using three DTDs. Additionally, the average MSE in this case is much higher than DTEC-Net. When the number of DTDs increases to two, the performance improves by 0.25dB and is already better than the SOTA method \cite{lyu2021u}. As we further increase the number of DTDs to three, the PSNR and SSIM increase 0.4dB and 0.003, respectively. The number of DTDs is finally set to three in a trade-off between computation and performance. To match different encoders and decoders and facilitate training, we also adjust the accept headers of Decoder1 to adapt to the sequence length determined by the different numbers of DTDs.

\begin{figure}
\centering
\includegraphics[width=0.95\textwidth]{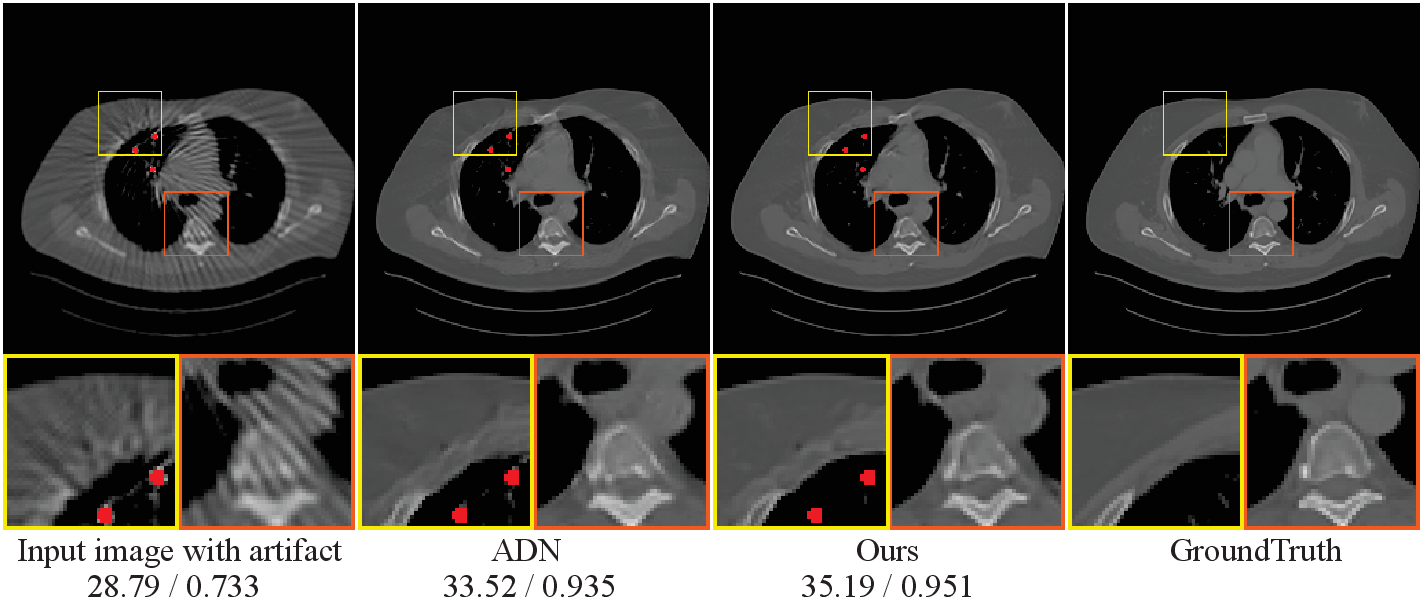}
\caption{Visual comparison(Metal implants are colored in \textcolor{red}{red}. The bottom values represent PSNR/SSIM). Our method has sharper edges and richer textures than ADN.}
\label{visulization1}
\end{figure}

\noindent
 \textbf{Only Transformer in DTD} Although the transformer can obtain better long-range correspondence than convolutions, it lacks the multiplexing of low-level information. For every DTD in DTEC-Net, we delete the second-order feature reuse pattern and only keep the lightweight transformer, the degraded version's results are 0.8dB lower than our DTEC-Net. At the same time, great instability appears in generative adversarial training. So, only using the transformer cannot achieve good results in reducing metal artifacts.

\noindent
\textbf{Removing SOD-MAR} Although SOD-MAR mainly helps by easing the burden of decoding as discussed in Section~\ref{sec:SOD-MAR}, it also has a performance gain compared to first-order disentanglement. We delete the SOD-MAR in DTEC-Net and let $x_h$ be the unique feature decoded by Decoder1. The Performance is 0.2dB lower than our DTEC-Net, while MSE increases by 1.47.

\subsection{Comparison to State-of-the-art (SOTA)}

For a fair comparison, we mainly compare with SOTA methods under unsupervised settings: ADN \cite{liao2019adn}, U-DuDoNet \cite{lyu2021u}, RCN \cite{zhao2020unsupervised}, and CycleGAN \cite{zhu2017unpaired}. We also compare with the traditional method LI \cite{kalender1987reduction} and classical supervised method CNNMAR \cite{zhang2018convolutional}. The quantitative results of ADN, CycleGAN, CNNMAR and LI are taken from \cite{liao2019adn}, the results of U-DuDoNet and RCN are taken from \cite{lyu2021u}. Because ADN has open-source code, we run their code for qualitative results. 

\begin{table}[t]
\begin{center}
\caption{\centering Quantitative comparison of different methods on Synthesized DeepLesion. The best results are marked \textcolor{red}{red}, the second best results are marked \textcolor{blue}{blue}.}
\label{tab2}
\setlength{\tabcolsep}{5mm}{
\begin{tabular}{ccll}
\toprule
Method Classification &  Method & PSNR$\uparrow$ & SSIM$\uparrow$ \\
\midrule
Conventional & LI\textsuperscript{\cite{kalender1987reduction}} & 32.00\textsuperscript{\cite{liao2019adn}} & 0.910\textsuperscript{\cite{liao2019adn}}\\
Supervised &  CNNMAR\textsuperscript{\cite{zhang2018convolutional}} & 32.50\textsuperscript{\cite{liao2019adn}} & 0.914\textsuperscript{\cite{liao2019adn}}\\
Unsupervised & CycleGAN\textsuperscript{\cite{zhu2017unpaired}} & 30.80\textsuperscript{\cite{liao2019adn}} & 0.729\textsuperscript{\cite{liao2019adn}}\\
Unsupervised & RCN\textsuperscript{\cite{zhao2020unsupervised}} & 32.98\textsuperscript{\cite{lyu2021u}} & 0.918\textsuperscript{\cite{lyu2021u}}\\
Unsupervised & ADN\textsuperscript{\cite{liao2019adn}} & 33.60\textsuperscript{\cite{liao2019adn}} & 0.924\textsuperscript{\cite{liao2019adn}}\\
Unsupervised & U-DuDoNet\textsuperscript{\cite{lyu2021u}} & \textcolor{blue}{34.54}\textsuperscript{\cite{lyu2021u}} & \textcolor{blue}{0.934}\textsuperscript{\cite{lyu2021u}}\\
Unsupervised & DTEC-Net(Ours) & \textcolor{red}{35.11} & \textcolor{red}{0.941}\\
\bottomrule
\end{tabular}
}
\end{center}
\end{table}

\noindent
\textbf{Quantitative Results}  As shown in Table~\ref{tab2}. For the Synthesized DeepLesion Dataset, our method has the highest PSNR and SSIM value and outperforms the baseline ADN by 1.51dB in PSNR and 0.017 in SSIM. At the same time, it also exceeds the SOTA method U-DuDoNet by 0.57dB. For the Real Clinic Dataset, the numerical results can’t be calculated because the ground truth does not exist. We will present the qualitative results in the appendix. Furthermore, as our work is single-domain based, it has the potential to be easily applied in clinical practice.

\noindent
\textbf{Qualitative Results} A visual comparison is shown in Fig.~\ref{visulization1}. Our method not only reduces artifacts to a large extent, but also has sharper edges and richer textures than the compared method. More results are shown in the appendix.

\section{Conclusion}
In this paper, we proposed a Dense Transformer based Enhanced Coding Network (DTEC-Net) for unsupervised metal-artifact reduction. In DTEC-Net, we developed a Hierarchical Disentangling Encoder (HDE) to represent long-range correspondence and produce an enhanced coding sequence. By using this sequence, the DTEC-Net can better recover low-level characteristics. In addition, to decrease the burden of decoding, we specifically design a Second-order Disentanglement for MA Reduction (SOD-MAR) to finish the sequence decomposition. The extensive quantitative and qualitative experiments demonstrate our DTEC-Net’s effectiveness and show it outperforms other SOTA methods.

\subsubsection{Acknowledgements.} This research work was undertaken in the context of Horizon 2020 MSCA ETN project “xCTing” (Project ID: 956172). 

\bibliographystyle{splncs04}
\bibliography{mybibliography}

\hspace*{\fill}
\\
\hspace*{\fill}
\section*{Appendix}
\begin{figure}[]
\centering
\includegraphics[width=\textwidth]{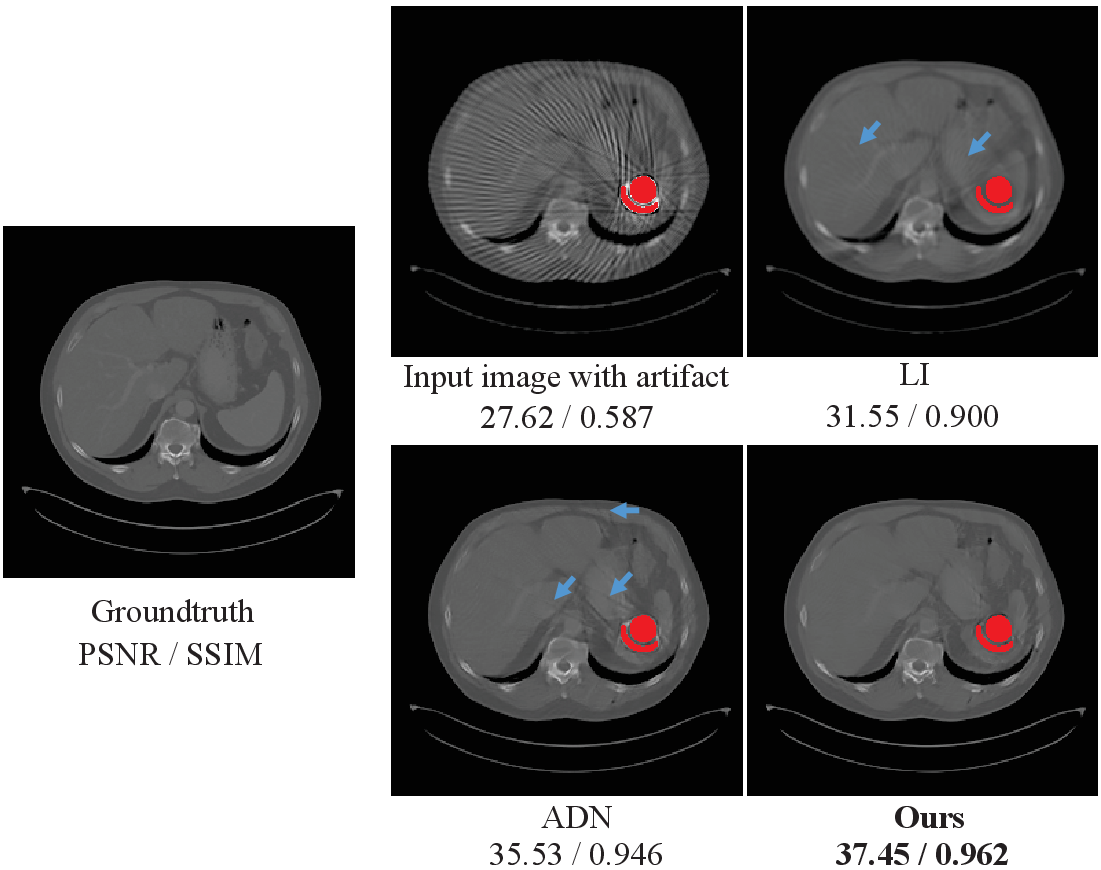}
\caption{Visual comparison on Synthesized DeepLesion Dataset. Metal implants are coloured  \textcolor{red}{red} . \textcolor{blue}{Blue} arrows indicate obvious artifacts that cannot be removed by comparison methods. The result shows our methods can greatly
reduce metal artifacts while restoring richer texture details than ADN and LI, which are cited in the main body of the paper.}
\label{fulu_1}
\end{figure}


\begin{figure}[t]
\centering
\includegraphics[width=\textwidth]{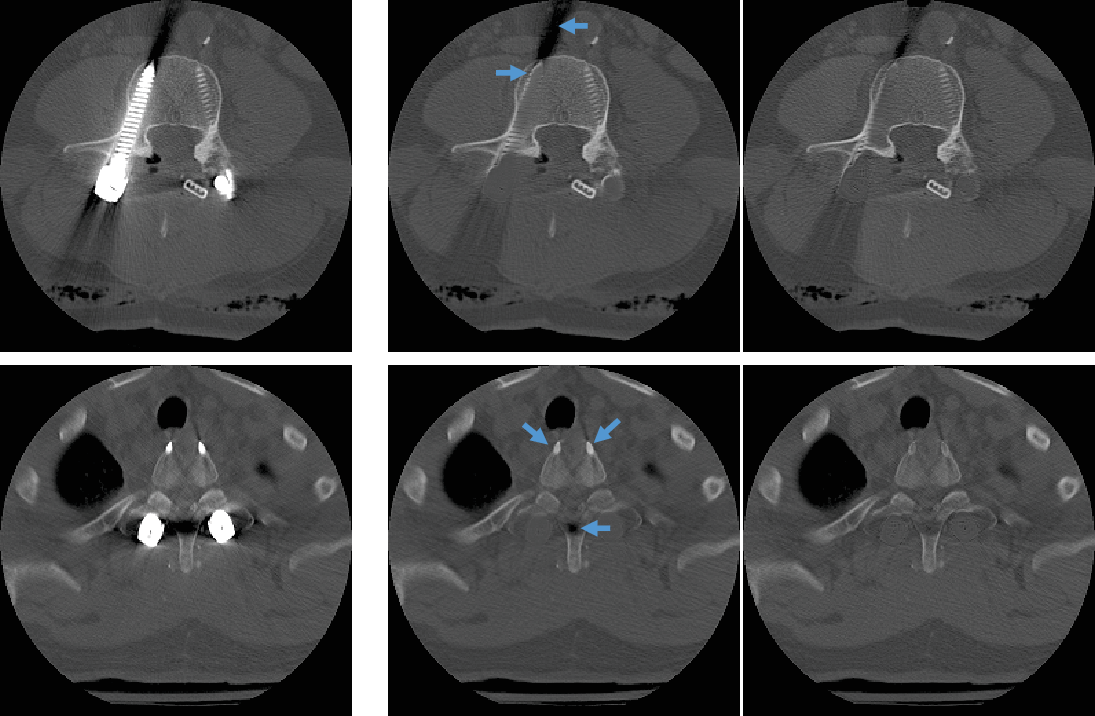}
\includegraphics[width=\textwidth]{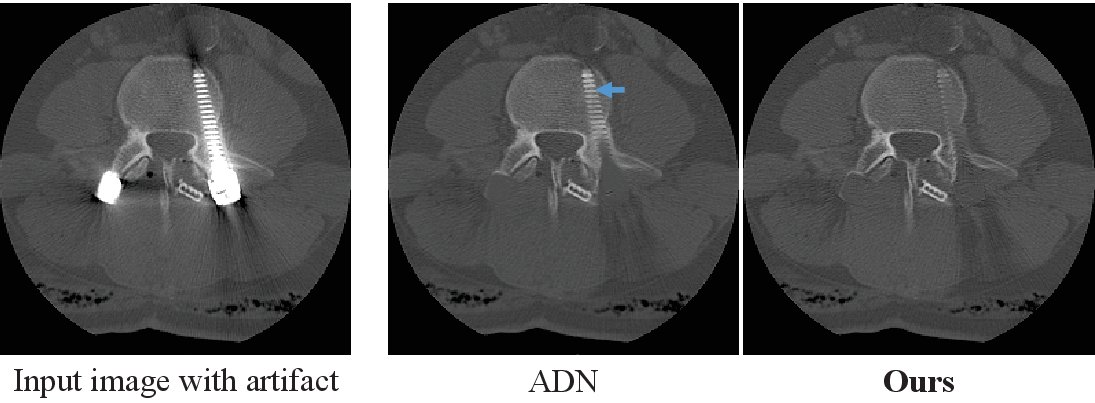}
\caption{Visual comparison on Real Clinic Dataset(SpineWeb). \textcolor{blue}{Blue} arrows indicate obvious artifacts that cannot be removed by comparison methods. Because the real clinic data set doesn't have ground-truth without artifacts, the numerical results of PSNR/SSIM can't be calculated. The result shows that our methods can better reduce the black band and bright area caused by metal plants compared to ADN.}
\label{fulu_2}
\end{figure}

\end{document}